# Multi-Persona Debate System for Automated Scientific Hypothesis Generation

Jaeha Oh[1,2], Byungchan Kim[1,2], Ju Li[3,4,*], Yang Jeong Park[5,6,*], Jin-Sung Park[1,2*]

[1]Department of Materials Science & Engineering, Ajou University, Suwon 16499, Republic of Korea

[2]Department of Energy Systems Research, Ajou University, Suwon 16499, Republic of Korea

[3]Department of Nuclear Science and Engineering, Massachusetts Institute of Technology, Cambridge, MA 02139, USA

[4]Department of Materials Science and Engineering, Massachusetts Institute of Technology, Cambridge, MA 02139, USA

[5]Department of Materials Science and Engineering, Ulsan National Institute of Science and Technology, Ulsan 44919, Republic of Korea

[6]Graduate School of Artificial Intelligence, Ulsan National Institute of Science and Technology, Ulsan 44919, Republic of Korea

email: liju@mit.edu, parkyj@unist.ac.kr, jinsung@ajou.ac.kr

## Abstract

Modern scientific discovery is bottlenecked not by data scarcity, but by the inability to synthesize fragmented knowledge into actionable hypotheses. This challenge is especially acute in battery materials research, where electrochemical performance, interfacial behavior, and manufacturing feasibility must be optimized simultaneously. Here, we present the Multi-Persona Debate System (MPDS), a literature-grounded framework for automated scientific hypothesis generation that combines literature retrieval, long-context large language model reasoning, corpus-driven persona induction, and structured multi-agent debate. MPDS constructs literature snapshots of up to 500 papers, grounds agents in role-specific evidence pools, and conducts a three-round citation-aware debate followed by moderator synthesis, enabling negotiation between personas while preserving evidence traceability. We evaluate MPDS using a temporally controlled protocol excluding direct access to target papers, including two held-out battery-materials case studies and a blinded comparison across 30 matched cases. In sodium-ion anode and all-solid-state battery cathode design tasks, MPDS recovered design logics aligned with experimentally validated solution spaces and generated more mechanistically explicit, process-aware proposals than simpler baselines. To assess the impact of personas and debate, we introduce Integrative Hypothesis Quality scoring. In ablation studies, MPDS achieved the highest mean score among five conditions, with its largest advantage in cross-perspective integration. A laboratory follow-up suggests utility as a diagnostic aid for identifying practical bottlenecks in workflows. These results indicate that structured debate over literature snapshots improves hypothesis formation under coupled engineering constraints and provides a reusable workflow for text-intensive scientific discovery.

## Introduction

The challenge of modern science is no longer information, but the capacity to integrate it. Recent analysis reveals that between 2016 and 2022, the total number of articles indexed in Scopus and Web of Science increased from approximately 1.92 million to 2.82 million, corresponding to roughly 47% growth over six years [1]. As this literature expands, individual researchers face a growing burden of knowledge, making broad synthesis harder and specialization more likely [2,3]. Information overload degrades decision quality and increases the risk that relevant evidence remains dispersed across disconnected literatures, leaving potentially valuable cross-domain links unnoticed [4,5]. This challenge is especially acute in battery materials science, where progress depends on jointly considering electrochemical performance, materials selection, interfaces, processing routes, manufacturability, and system-level constraints rather than optimizing a single variable in isolation [6,7]. Recent science-of-science research has shown that large-scale bibliometric datasets can reveal patterns in how discovery unfolds, motivating scalable computational systems for literature synthesis across large bodies of scientific work [8].

However, current AI approaches for hypothesis generation remain limited in reliability and evidence traceability in scientific settings. Large Language Models (LLMs) have emerged as promising tools for knowledge synthesis, but standard implementations (e.g., zero-shot prompting of GPT-4 [9]) face critical limitations when applied to scientific reasoning. Structured reasoning approaches such as Chain-of-Thought and self-consistency can improve reliability in complex tasks [10,11]. Despite these advances, standard LLM pipelines still suffer from three coupled failure modes: hallucinated or weakly grounded statements, single-perspective reasoning that underrepresents engineering trade-offs, and insufficient evidence traceability for experimental planning. These limitations reduce reliability when generated hypotheses must be translated into actionable laboratory decisions [12,13].

Recent studies suggest that multi-agent debate can improve reasoning and factuality in generic tasks such as mathematical and strategic reasoning [14]. Persona-driven collaboration has likewise been shown to enrich ideation by broadening perspective diversity, increasing idea depth, and improving cross-domain coverage in brainstorming and argument generation [15,16]. More recently, multi-agent LLM systems have begun to extend collaborative generation beyond generic reasoning and brainstorming to selected scientific workflows and AI-assisted scientific idea generation [17,18]. However, it remains comparatively underexplored whether such methods can robustly support hypothesis generation in battery materials research, where electrochemical, interfacial, and manufacturing variables interact in a highly coupled, nonlinear design space [6,7].

Here we propose the Multi-Persona Debate System (MPDS), a framework that integrates API-driven large-scale literature access, long-context LLM reasoning, and multi-persona debate to generate hypotheses through explicit cross-validation of constraints and trade-offs. MPDS uses OpenAlex [19] API-driven literature retrieval over a corpus spanning hundreds of millions of scholarly records and aggregates hundreds of abstracts (up to 500 papers in this study) into a long-context evidence pool for downstream reasoning. Rather than relying on conventional top-k chunk retrieval in vector RAG, this design enables agents to reason over relationships across papers within a long-context window [20,21].

From the retrieved literature pool, MPDS induces complementary personas, lets them debate the research problem, and uses a moderator agent to synthesize the exchange into a final literature-grounded proposal.

To evaluate MPDS, we therefore adopted a time-locked validation strategy in which fixed literature snapshots assembled through OpenAlex were used to test whether literature-grounded multi-agent reasoning can generate evidence-traceable and experimentally relevant hypotheses under temporally controlled conditions. This validation proceeded along two complementary axes: first, two held-out battery-materials case studies grounded in Park et al. (2017) and Kim et al. (2025) [22,23], which allowed us to examine whether MPDS could recover the governing design logic of experimentally validated problems without direct access to the reference papers; and second, a custom-rubric-based quantitative evaluation across 30 matched battery-materials cases, designed to assess evidence traceability [12,13], grounded hypothesis construction, and the comparative contribution of debate-based reasoning relative to simpler baselines [14].

Using this evaluation framework, we find that MPDS can recover literature-grounded design spaces that are well aligned with experimentally validated battery-materials solutions and can produce more mechanistically explicit, process-aware proposals than simpler baselines. In a blinded comparison, introducing persona induction or debate each improved performance beyond the evidence-only setting, and the combination of the two achieved the strongest overall results. A laboratory follow-up further suggests potential utility as a diagnostic aid for narrowing practical experimental bottlenecks. These results support MPDS as a reusable framework for literature-grounded hypothesis formation under coupled design constraints.

## Methodology

MPDS was designed for battery materials problems in which materials chemistry, interfacial behavior, and process feasibility must be considered together. On this basis, MPDS operates through four linked stages (Fig. 1). First, the system retrieves topic-relevant papers through the OpenAlex API and consolidates their abstracts into a long-context evidence pool, so that hypothesis generation begins from a broad but bounded body of prior work rather than from a few isolated passages. Second, this literature is used to instantiate domain-specific personas, a step that turns the same evidence base into complementary interpretive viewpoints and helps expose trade-offs that a single perspective might overlook. Third, the personas debate the user-defined research topic over three rounds, during which competing proposals are iteratively critiqued and revised so that hidden failure modes, feasibility gaps, and coupled design constraints become more explicit. Fourth, a moderator synthesizes the two debate outputs into a final recommendation, turning the debate into a single actionable proposal while preserving links to the supporting evidence.

### Literature retrieval and corpus construction

To mitigate the “frozen knowledge” problem of standard LLMs, MPDS utilizes the OpenAlex API, a fully open index of scholarly works, authors, venues, institutions, and concepts, as its dynamic knowledge base [19]. For each topic, MPDS applies temporal filtering, reconstructs OpenAlex inverted-index abstracts into readable text, and aggregates up to 500 high-relevance papers into a time-locked literature snapshot for downstream reasoning. This retrieval stage preserves breadth while keeping processing costs tractable.

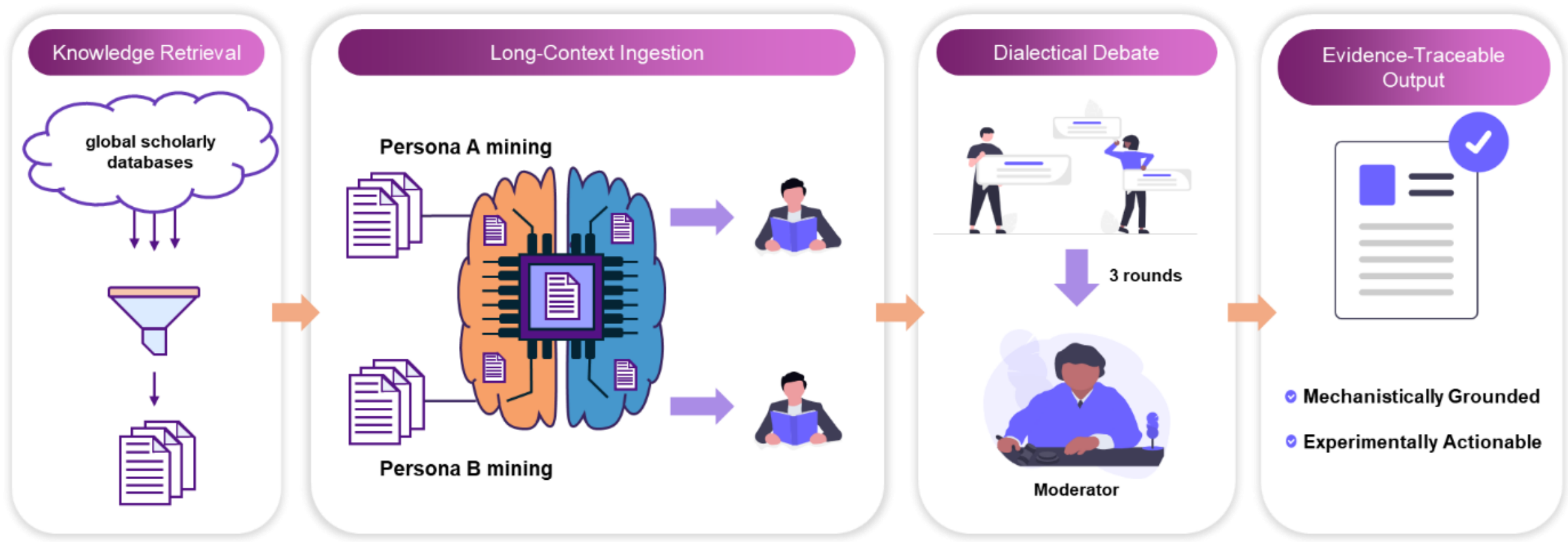

**Figure 1 | Architecture of the Multi-Persona Debate System (MPDS).** The pipeline consists of four stages: (1) *Knowledge Retrieval*, which filters and curates documents from global scholarly databases; (2) *Long-Context Ingestion*, in which two independent AI personas (A and B) mine the corpus in parallel; (3) *Dialectical Debate*, where the personas engage in three rounds of moderated adversarial exchange; and (4) *Evidence-Traceable Output*, producing a structured report that is mechanistically grounded and experimentally actionable. Illustrations adapted from unDraw (undraw.co), used under the unDraw License.

## Long-context ingestion and agent instantiation

Unlike top-k passage retrieval [20] in vector RAG, this module performs corpus-level long-context ingestion so agents can reason over inter-paper dependencies rather than isolated fragments. MPDS grounds each debater in a role-specific, time-locked long-context evidence pool, enabling within-snapshot cross-paper reasoning while preserving complementary perspectives across the debate. This design reduces context fragmentation and better supports cross-paper synthesis across the retrieved literature. More broadly, recent benchmarking suggests that, when sufficiently resourced, long-context LLMs can outperform retrieval-based chunking approaches in average performance across a range of long-context, query-based tasks [21]. However, long-context models can exhibit positional bias and may underutilize evidence located in the middle of very long inputs, so corpus ordering and summarization strategy remain important [24].

To simulate interdisciplinary scientific discourse, MPDS instantiates distinct AI agents using Gemini 2.5 Pro [25], which was selected for its reasoning capability over long contexts. For reproducibility, the generation temperature was fixed at 0.5 across all conditions to reduce stochastic variation while preserving sufficient flexibility for evidence-grounded hypothesis synthesis.

Before debate, MPDS performs corpus-driven persona induction by mining each retrieved evidence pool for domain-specific keyword signatures, recurrent problem-solving patterns, and characteristic reasoning priorities. This step assigns each agent a domain-aligned viewpoint, such as materials-focused or process-focused reasoning, and helps distribute attention across complementary constraints within the same research problem. For computational tractability, persona induction was performed on a fixed-length 70,000-character excerpt from each role-specific evidence pool, whereas the subsequent debate used the full retrieved evidence pool. MPDS supports multiple complementary personas depending on task complexity; in the present experiments, we use two debaters (materials-focused and process-focused) plus one moderator for final synthesis. This configuration covers both mechanism design and realization constraints while keeping the chain-of-debate compact and reproducible.

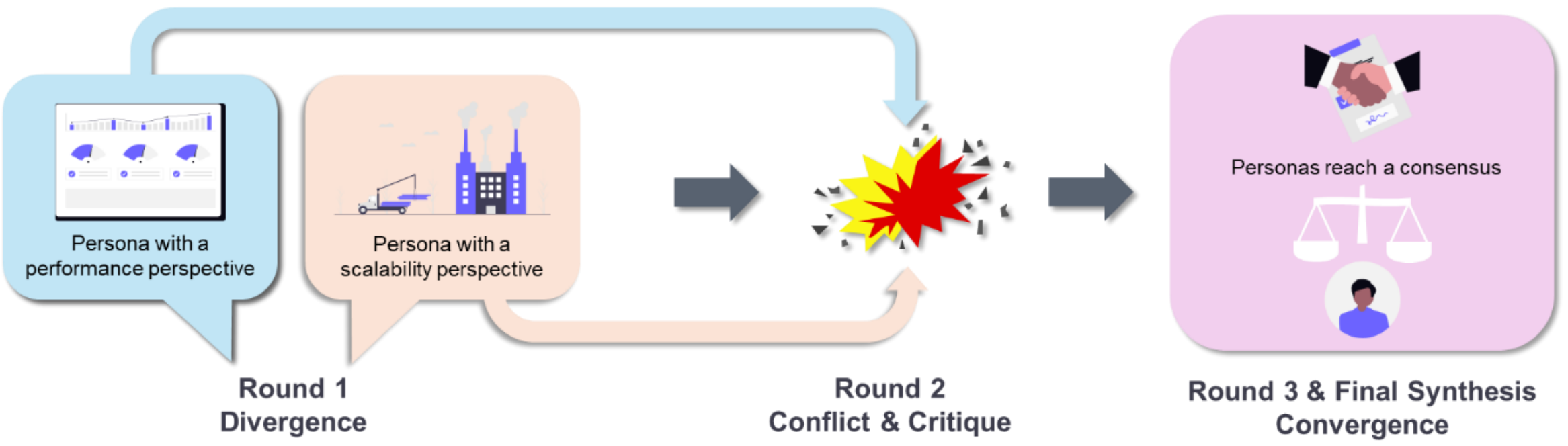

**Figure 2 | Three-Round Dialectical Debate and Moderator Synthesis Workflow.** In Round 1, the personas diverge by advancing initial proposals from distinct priorities, such as performance and scalability. In Round 2, they enter conflict and critique, surfacing hidden failure modes, unresolved trade-offs, and feasibility gaps. By Round 3, the exchange converges toward a compromise solution, which is then consolidated by a moderator into a final consensus plan with explicit evidence traceability. Illustrations adapted from unDraw (undraw.co), used under the unDraw License.

## Structured co-design debate and synthesis

This adversarial but citation-aware exchange is designed to prevent premature convergence, maintain provenance, and progressively sharpen the design toward experimentally actionable constraints. The core reasoning engine therefore implements a three-iteration, citation-grounded co-design debate in which each turn couples material-side proposals with process-side realization constraints. Fig. 2 summarizes this iterative debate workflow and the post-debate synthesis stage. Across the three iterations, each agent critiques the opponent's latest proposal by identifying hidden failure modes, unresolved trade-offs, and feasibility gaps, and then submits a revised material–process package grounded in cited evidence from the ingested corpus. After the third iteration, a moderator agent produces a final compromise plan by integrating compatible components from both reasoning streams while preserving traceable links to the supporting evidence.

## Time-Locked Validation Protocol

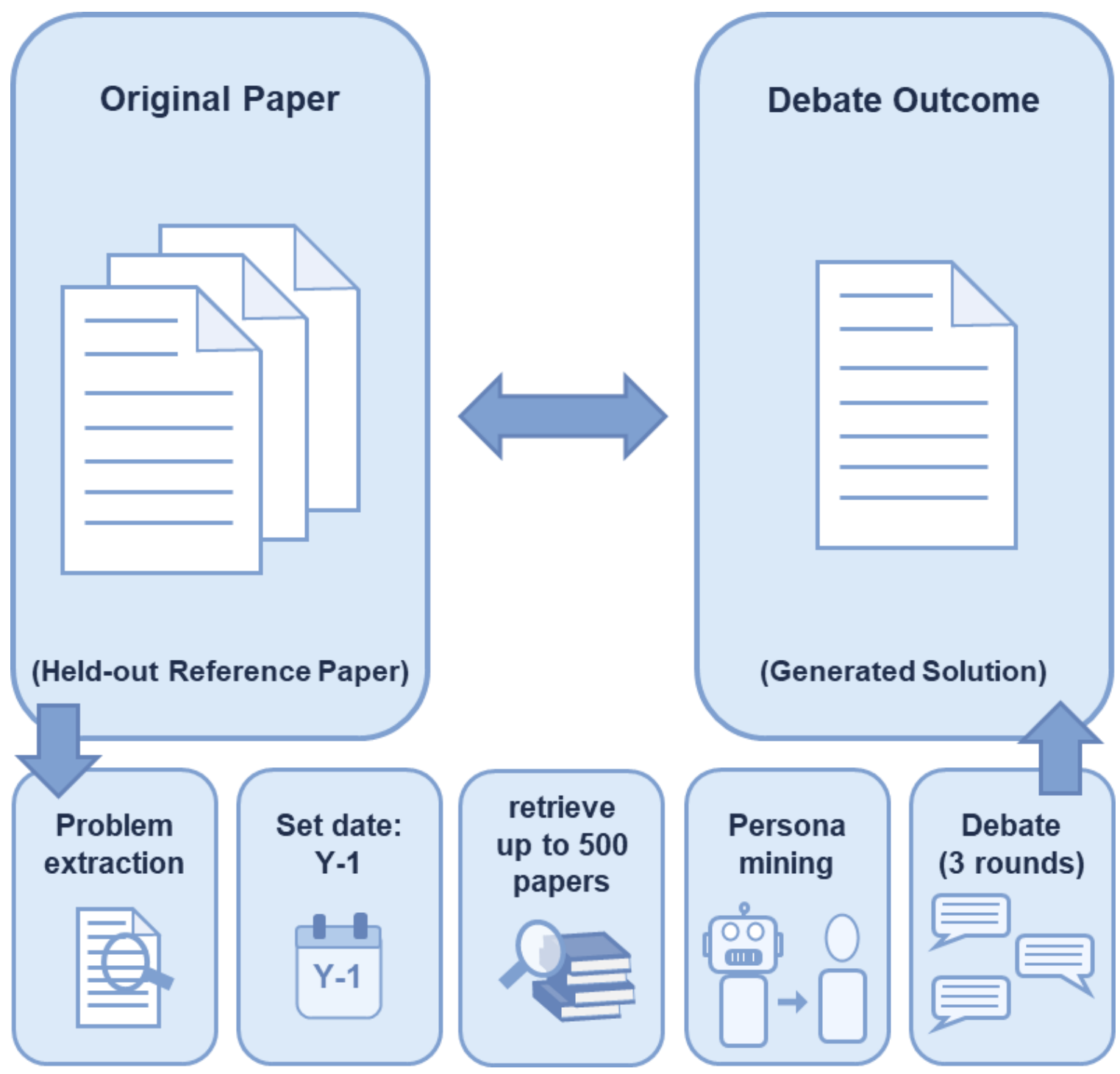

**Figure 3 | Time-Locked Validation Framework for MPDS.** For each benchmark case, the original reference paper is held out and transformed into a problem statement rather than being shown directly to the model. A temporal cutoff is then set to one year before publication (Y−1), after which MPDS retrieves up to 500 papers, mines complementary personas from the resulting snapshot, and runs a three-round debate to generate a literature-grounded solution. The debate outcome is finally compared post hoc with the held-out ground-truth paper to test whether MPDS can recover the governing design logic under temporally restricted evidence.

To assess MPDS under a time-locked retrieval setting, Fig. 3 summarizes the validation protocol used in both case studies. Rather than allowing MPDS to read the reference paper and reproduce the published solution directly, we held out the reference paper itself and provided only a manually specified design problem derived from it. For each case, the problem specification included the central bottleneck, coupled constraints, and target application. To mitigate the risk of data leakage and facilitate a rigorous evaluation of the system's de novo reasoning capacity, the information cutoff was set to one year before the paper's publication year (Y-1). This time lock excluded the reference paper itself and the same-year literature from the MPDS knowledge base.

Using this date anchor, MPDS assembled a time-locked literature pool from OpenAlex [19]. In the benchmarked case studies, these retrieval pools were stored as pre-fetched snapshots to support deterministic replay and fair comparison, even though the same workflow can operate with live API retrieval in deployment.

Up to 500 papers were retrieved for each keyword-defined snapshot, and the resulting abstract sets served as fixed long-context evidence pools throughout the benchmark. Under the same date constraint, MPDS was then executed with its standard pipeline of persona induction, three-round debate, and moderator synthesis. To preserve provenance, each response was validated against indexed evidence identifiers from the corresponding role-specific snapshot, and the final synthesis was required to incorporate evidence from both debate sides.

The generated final solution was then compared post hoc with the held-out reference paper. This comparison was intended as a directional validation rather than an exact reconstruction test: the central question was whether MPDS could recover the governing design logic, coupled constraints, and plausible resolution pathway under temporally restricted retrieval, not whether the backbone model had zero prior parametric exposure to the held-out paper.

## Case Studies: MPDS Application in Battery Materials Science

Following the protocol in Fig. 3, we next evaluate two held-out battery-materials case studies to test whether MPDS can recover a literature-grounded design space that addresses the same coupled bottlenecks without direct access to the held-out reference paper. These two cases were chosen to represent distinct forms of constraint-coupled reasoning for which multi-perspective debate is especially useful.

Case Study 1 (sodium-ion battery anode design) examines a materials–manufacturing tension: how to achieve cycling-stable, high-capacity anodes under large volume expansion while retaining scalable particle-level manufacturability. The held-out reference for this case is Park et al. (2017) [22]. MPDS uses two complementary personas here: one focused on sodium-ion anode chemistry and volume-expansion mitigation, and the other on particle-level engineering, including architectural design, densification/porosity trade-offs, and scalable processing constraints, followed by moderator-guided synthesis.

Case Study 2 (all-solid-state battery cathode composite design) targets interface optimization: how to enhance ionic conduction through cathode active material (CAM)–solid electrolyte (SE) contact while preserving electronic access for charge transfer. The held-out reference for this case is Kim et al. (2025), which introduced the half-covered glitter-cake active material (AM)@SE composite cathode through mechanofusion and particle-size engineering [23]. MPDS again uses two complementary personas: a sulfide-solid-electrolyte persona (interfacial ion transport and coating/shell geometry) and an all-solid-state-battery-manufacturing persona (dry-process feasibility, scale-up constraints, and robustness), followed by moderator synthesis.

Accordingly, Tables 1 and 2 summarize the two complementary comparisons for Case Study 1, and Tables 3 and 4 do the same for Case Study 2. Within each case, one table compares the final MPDS output with the held-out reference paper, whereas the other compares MPDS with a Raw LLM baseline in terms of mechanistic depth. The Raw LLM baseline used the same backbone model (Gemini 2.5 Pro) but operated as a single-pass, parametric-only response without multi-persona debate, external corpus reasoning, or moderator synthesis.

## Case Study 1: Sodium-Ion Battery Anode with Template-Free Scalable Synthesis

## Problem Statement and Initial Observations

This case concerns metal chalcogenide (sulfide/selenide) anodes for sodium-ion batteries (SIBs), where scalable synthesis and volume buffering must be solved together. Sodium-ion batteries have attracted interest as cost-effective options for large-scale energy storage because sodium is abundant and compatible with established battery-engineering paradigms [26]. While metal chalcogenide anodes offer high theoretical capacities (500–800 mAh $g^{-1}$) when applied as anodes for SIBs, they face two coupled engineering bottlenecks. First, synthesis scalability remains limited: common hydro/solvothermal routes are time-intensive, liquid-phase batch processes that often yield non-uniform nanostructures and do not readily scale beyond laboratory quantities. Second, volume-expansion tolerance is critical: SIB cycling can induce large volume changes during Na insertion/extraction, motivating porous or void-containing architectures. However, many prior solutions rely on hard/soft templating (e.g., polymer spheres or silica) to create internal void space, and template removal increases cost and process complexity in ways that are difficult to reconcile with commercial-scale production. The central challenge, therefore, is to design a scalable, template-free synthesis route that yields uniform micrometer-sized spheres while incorporating internal porosity to buffer volume expansion.

## MPDS Debate Summary: Integrated Synthesis

In Round 1, the sodium-ion-anode expert framed the task as a stability–scalability conflict and argued that sodium-specific conversion chemistry demands porous buffering architectures that mitigate large volume changes without the use of hard-to-remove templates. In Round 2, the particle-engineering expert emphasized reproducibility, phase purity, and process-controllability constraints—highlighting that scalable routes must avoid sensitive, failure-prone steps—while the anode expert stressed that practical SIB deployment should preserve low-cost robustness and electrolyte accessibility rather than pursuing idealized uniformity alone. By Round 3, the exchange converged on co-optimizing morphology, transport pathways, and controllable processing knobs, shifting from single-metric tuning to a coupled design strategy suitable for pilot-scale translation. The final compromise proposed a one-pot, scalable particle-level synthesis route, targeting porous micro-assembled spheres that preserve tap-density practicality while providing internal strain buffering and manufacturing stability (Supplementary Fig. 1).

## Comparison between the held-out reference and MPDS (Table 1)

We compare the MPDS final compromise in this case against Park et al. (2017), which synthesized $MoS_2$–$Ni_9S_8$ microspheres via dextrin-assisted spray drying, where dextrin acted as both a carbon source and a nanovoid-forming agent [22]. The reference paper validated template-free spray-dried Mo/Ni chalcogenide microspheres with uniformly distributed empty nanovoids. MPDS did not reproduce that exact implementation route; instead, it converged on a closely related design space centered on hierarchical spray-dried microspheres, distributed internal buffering, and conductive-network integration. Table 1 therefore compares the implementation validated in the held-out reference paper with the MPDS design route at the level of design objective, particle architecture, buffering logic, conductivity strategy, and process route.

### Comparison in mechanistic depth against a Raw LLM baseline (Table 2)

Compared with the Raw LLM baseline, which prioritized a carbon-coated yolk-shell motif, MPDS shifted the design toward a spray-dried secondary-particle architecture that explicitly incorporates tap density, process scalability, and composite-level transport continuity. Table 3 shows that this difference is not merely limited to morphology. The Raw LLM response treated volume buffering mainly as a local structural problem, emphasizing hollow or yolk-shell protection but leaving electrode-level packing, conductive-network continuity, and process translation comparatively underdeveloped. By contrast, MPDS reframed the problem at the particle and electrode levels simultaneously: it retained internal strain-buffering logic, but embedded that logic within a spray-dried, hierarchically assembled secondary-particle route that is more compatible with scalable powder processing.

## Case Study 2: All-Solid-State Battery Cathode Composite

### Problem Statement and Initial Observations

The second case concerns high-loading (>85 wt%) composite cathodes for sulfide-based all-solid-state batteries (ASSBs). Despite the strong safety and potential for high energy density, ASSBs remain constrained by unresolved interfacial and process-integration challenges that complicate practical scale-up [27]. The challenge is to integrate three components with fundamentally conflicting electrochemical and thermal/process constraints: (1) CAM, which serves as the host for lithium storage and redox reactions; (2) SE, which provides ionic conduction but is electronically insulating; and (3) an electronically conductive network, depending on how the composite electrode is designed. Notably, in the Kim et al. (2025) reference, carbon is not treated as the primary design axis; the central design focus is CAM–SE geometry and coating coverage, with electronic access governed largely by contact architecture (e.g., CAM–CAM accessibility and partial SE coverage). The core paradox is that complete SE coating ensures excellent ionic contact but blocks electronic access, whereas minimizing SE coating preserves electronic pathways but produces fragmented ionic networks at >85 wt% loadings. The central design challenge is therefore to simultaneously maximize ionic conductivity (SE–CAM contact) and maintain electronic accessibility (electron-conductive pathways governed by contact geometry and, if used, conductive additives) within a >85 wt% CAM composite cathode while preserving manufacturing simplicity and material stability.

### MPDS Debate Summary: Two-Persona Integrated Solution

In Round 1, the sulfide-solid-electrolyte persona framed the task as a geometric allocation problem, arguing that limited LPSCl must be split between particle-surface coverage and interstitial filling to preserve ionic continuity at >85 wt% active material. The manufacturing-oriented persona emphasized process robustness, preservation of the fixed polycrystalline NCM811 secondary-particle platform, and the need to avoid overly aggressive processing that could damage either NCM811 or LPSCl. Through critique and revision, the debate converged on an explicit mechanofusion-based allocation rule within the fixed NCM811/LPSCl/conductive-additive composite: a minority nano-LPSCl fraction for partial NCM811 surface coverage, a majority micron-scale LPSCl fraction for interstitial filling, and a sequential low-energy mechanofusion route designed to preserve particle integrity while maintaining electronic accessibility and dense packing (Supplementary Fig. 2).

### Comparison between the held-out reference and MPDS (Table 3)

The resulting debate from MPDS is compared with the half-covered glitter-cake concept reported by our group in Kim et al. (2025), which was realized through mechanofusion and particle-size engineering [23]. The reference paper validated a half-covered glitter-cake cathode in which a thin mechanofused LPSCl shell and small solid-electrolyte fillers jointly balanced ionic and electronic transport at high active-material loading. As summarized in Table 2, MPDS remained within the same fixed NCM811 secondary-microsphere platform and the same high-loading design problem, but turned a broad trade-off into an allocation rule by recasting it as a limited-LPSCl allocation problem under mechanofusion.

Specifically, MPDS assigns a minority nano-LPSCl fraction to partial surface coverage in order to secure local ionic access, while a majority micron-scale LPSCl fraction is used for interstitial filling to build bulk ionic pathways without forming a fully insulating shell. At the same time, exposed NCM regions and high-aspect-ratio conductive additives preserve long-range electronic percolation. Table 2 indicates that MPDS does not move to a fundamentally different cathode concept; rather, it clarifies the transport, stabilization, and process logic of the same mechanofusion-based design space through incomplete coverage and a sequential low-energy route in which nano coverage is introduced first and micron filler is added second.

### Comparison in mechanistic depth against a Raw LLM baseline (Table 4)

Compared to the Raw LLM baseline, MPDS did not move to a different cathode concept; instead, it remained within the same design space and made the design logic more explicit and experimentally actionable. As summarized in Table 4, the Raw LLM baseline stayed at the level of a broad geometric comparison among surface-biased, interstitial-biased, and hybrid LPSCl allocation strategies, whereas MPDS translated this into a more concrete rule linking LPSCl allocation, conductive-network design, and mechanofusion sequence. MPDS also expanded the failure analysis by incorporating process- and interface-linked risks, including secondary-particle damage, SE-conductivity loss, and SE–carbon/interface instability, while clarifying the central trade-off by separating the roles of nano-scale surface coverage, micron-scale fillers, and preserving the electronic pathway.

## Further validation

### Integrative Hypothesis Quality (IHQ)-based quantitative evaluation

To test whether the qualitative advantages observed in the case studies generalized beyond the two held-out examples, we evaluated ablation studies on five system conditions (Fig. 4) - Raw LLM, Evidence-Only (EO), Evidence-Only Persona (EOP), Debate System (DS), and MPDS - across 30 matched battery-materials cases using blind IHQ scoring. EO denotes a single-pass baseline conditioned only on a merged long-context evidence snapshot. EOP denotes a single-pass baseline that adds evidence-derived persona induction to the same merged long-context snapshot, but does not include multi-round debate or moderator synthesis. DS denotes a persona-free debate baseline that preserves evidence-grounded multi-round exchange without persona induction. MPDS extends DS by adding role-specialized persona induction, whereas Raw LLM denotes a parametric-only single-pass baseline without external evidence. Across the five conditions, a matched case refers to the same held-out problem setting derived from a given reference paper and evaluated under the same temporal cutoff; for the evidence-enabled conditions, the corresponding fixed retrieved evidence snapshots were also held constant so that only the reasoning structure varied across systems. A summary of one illustrative debate case is provided in Supplementary Fig. 3. All final outputs were anonymized and re-scored under a fixed IHQ prompt using Gemini 2.5 Pro as the blind judge. Before scoring, the judge evaluated only the final output for each case, without being told which system had produced it or how it had been generated, under the same fixed IHQ prompt. The full list of the 30 benchmark cases and their source reference papers is provided in Supplementary Table 1.

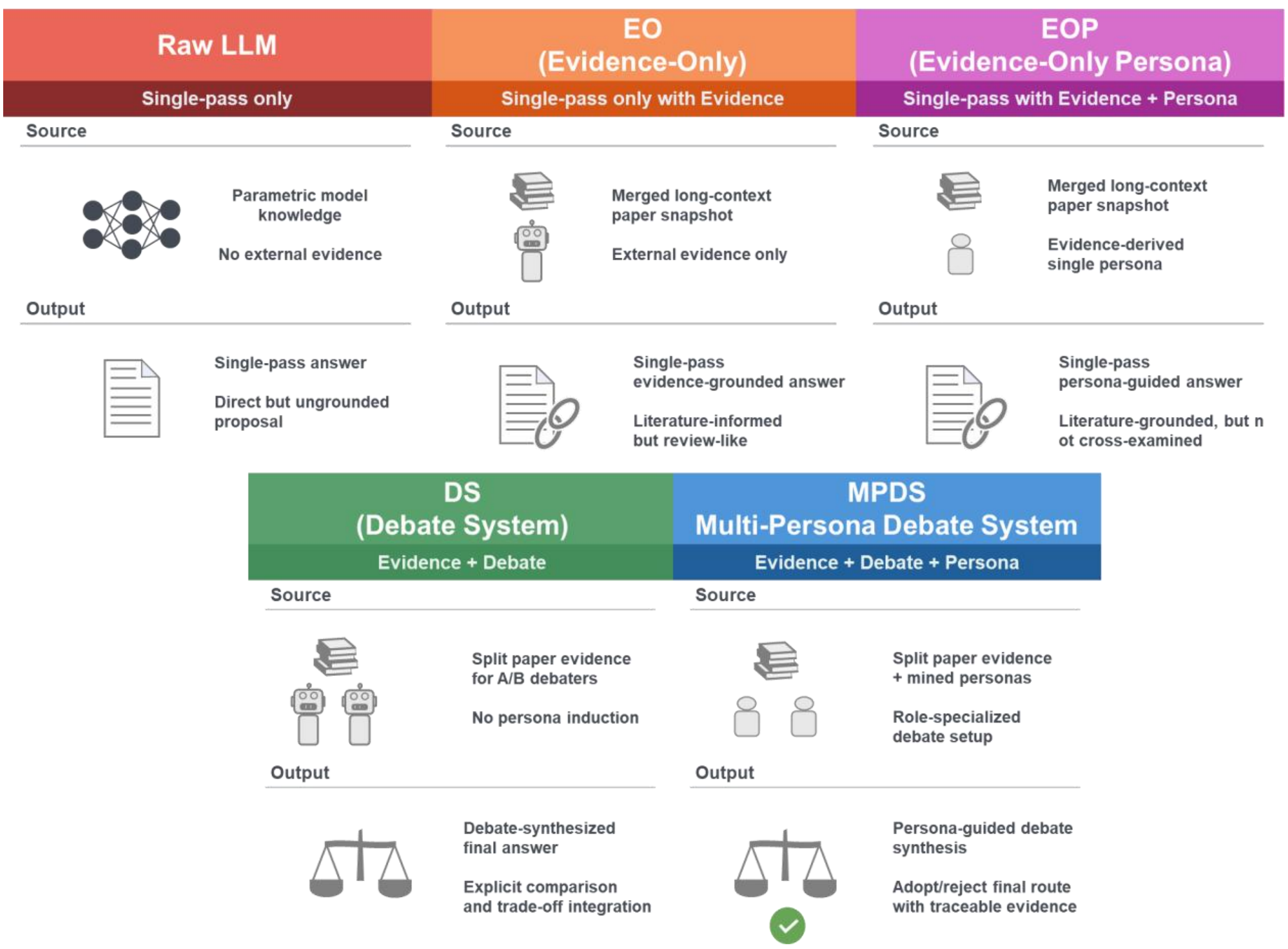

**Figure 4 | System Conditions Compared in This Study: Raw LLM, EO, EOP, DS, and MPDS.** Raw LLM generates a single-pass answer from parametric knowledge only, without external evidence. EO conditions the same backbone model on a merged long-context evidence snapshot but remains single-pass and non-personalized. EOP adds evidence-derived persona induction to the same merged snapshot while retaining single-pass generation. DS introduces evidence-grounded, multi-round debate with split evidence pools for two debaters but without persona induction. MPDS combines split evidence, role-specialized persona induction, multi-round debate, and moderator synthesis, yielding the most fully integrated and traceable final output.

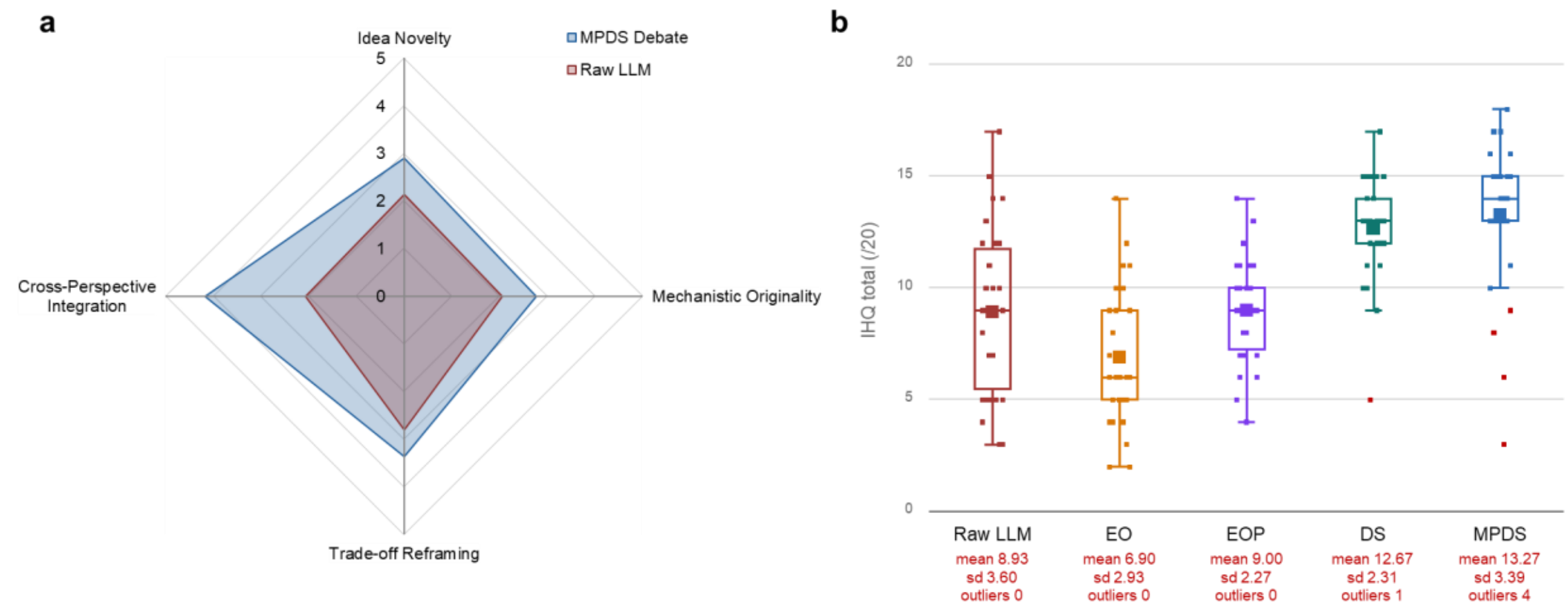

**Figure 5 | Blind Universal IHQ Comparison Across Raw LLM, EO, EOP, DS, and MPDS.** (**a**) Mean axis-wise IHQ profile comparing MPDS with the parametric-only Raw LLM baseline across Idea Novelty, Mechanistic Originality, Trade-off Reframing, and Cross-Perspective Integration. (**b**) Distribution of total IHQ scores across 30 matched battery-materials cases for all five system conditions. MPDS achieved the highest mean total score (13.27/20), followed by DS (12.67/20), EOP (9.00/20), Raw LLM (8.93/20), and EO (6.90/20), indicating that the strongest performance emerges when persona induction and evidence-grounded debate are combined.

The IHQ rubric scores each response on four dimensions - Idea Novelty, Mechanistic Originality, Trade-off Reframing, and Cross-Perspective Integration (Supplementary Fig. 4) - on a 0-5 scale, yielding a total score out of 20. Because LLM-based judges provide scalable and standardized comparative scoring but remain sensitive to evaluator configuration, prompt structure, and rubric wording, we used IHQ as a fixed within-study comparator rather than as an absolute or model-invariant measure of hypothesis quality. To reduce case-specific variance from any single judgment, all conditions were evaluated across 30 matched cases under one blinded judge protocol, and inference was based primarily on aggregate score patterns rather than on isolated case-level scores.

Fig. 5 summarizes the blind five-system comparison. Across the 30 completed cases, MPDS achieved the highest mean IHQ total score (13.27/20), followed by DS (12.67/20), EOP (9.00/20), Raw LLM (8.93/20), and EO (6.90/20). Relative to EO, both persona induction and evidence-grounded multi-round debate improved performance, and the highest score was obtained when the two were combined in MPDS. The EO-to-EOP improvement suggests that persona induction helps convert the same evidence pool into a more directed and mechanistically interpretable single-pass proposal, rather than leaving the output at the level of a more review-like evidence summary. The larger EO-to-DS improvement indicates that debate strengthens the explicit reconciliation of competing constraints through critique and revision. Most importantly, the EO-to-MPDS improvement shows that combining persona induction with debate yields the strongest overall performance, bringing together sharper framing with more explicit cross-perspective integration. The largest separation remained concentrated in cross-perspective integration, suggesting that the most effective outputs emerge when debate-driven integration is paired with persona-guided structuring.

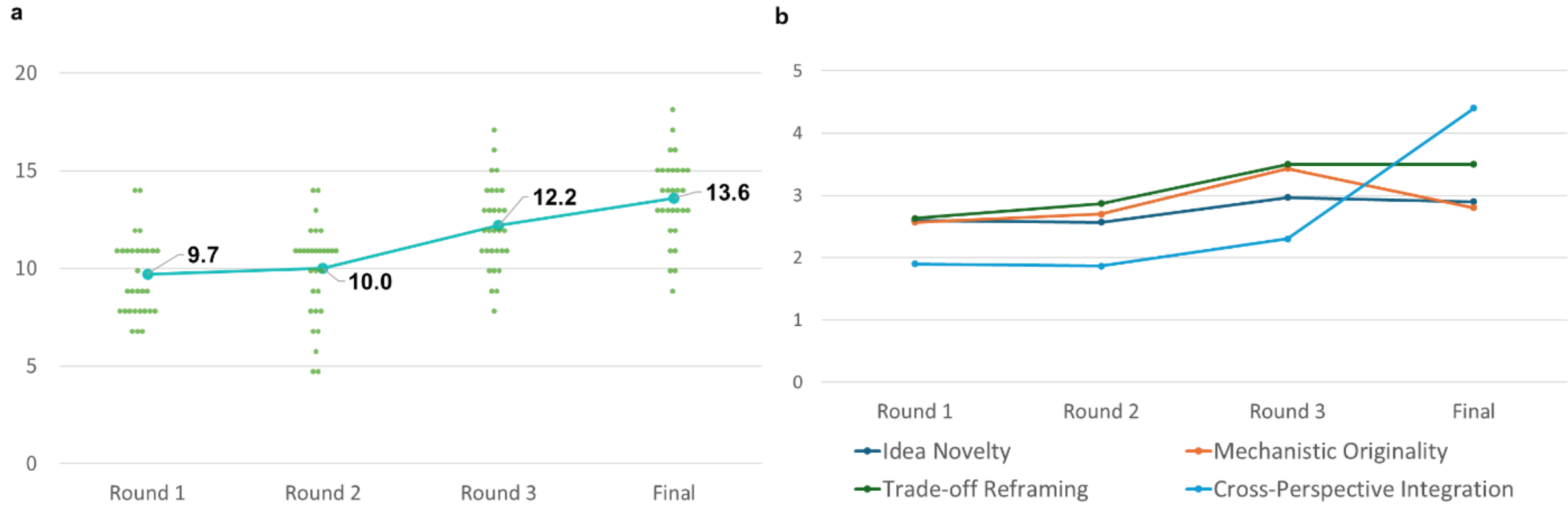

**Figure 6 | Stage-wise refinement trajectory of MPDS under Integrative Hypothesis Quality (IHQ). (a)** Mean total IHQ score across the MPDS chain-of-debate stages, from Round 1 through the final synthesis, showing a general increase in score over the debate trajectory. (**b)** Axis-wise mean IHQ scores across the same stages for Idea Novelty, Mechanistic Originality, Trade-off Reframing, and Cross-Perspective Integration. Later-stage outputs tended to score more highly overall, with the clearest improvement observed in cross-perspective integration. Because stage identity may remain partially inferable from text structure, these results should be interpreted as suggestive rather than definitive evidence of within-debate refinement.

To further examine whether refinement accumulates across debate rounds, we performed an additional blinded stage-wise re-scoring of MPDS chain-of-debate outputs. Fig. 6 shows that later-stage outputs tended to achieve higher IHQ scores, particularly in cross-perspective integration. Because some stage identity may remain partially inferable from text structure, however, these results should be interpreted as suggestive rather than definitive evidence of within-debate refinement.

## Post-debate laboratory follow-up

Fig. 7 reports a post-debate laboratory follow-up rather than a direct, step-by-step implementation of the final synthesized plan. The "Before" condition corresponds to a literature-guided initial experiment, whereas the "After" condition corresponds to a revised experiment after MPDS identified the likely bottleneck and suggested targeted process adjustments. In a laboratory follow-up experiment on the Prussian blue analogue (PBA) synthesis sequence conducted in our group, the debate output was used to diagnose the dominant failure source, identifying the washing step as the practical bottleneck governing reproducibility and material integrity. After isolating this root cause, the protocol was selectively adjusted by reducing washing intensity and solvent usage, including a shift from vacuum filtration to centrifugation and a reduction in wash volume from 600 mL to below 50 mL. These observations indicate that MPDS served primarily as a diagnostic reasoning aid rather than as a literal experimental protocol. In experimental practice, pinpointing the cause of repeated failure can require substantial time because multiple coupled variables must be disentangled through iterative trial-and-error and literature review; in this setting, MPDS may help reduce diagnostic latency by narrowing plausible failure sources and suggesting a more focused corrective pathway.

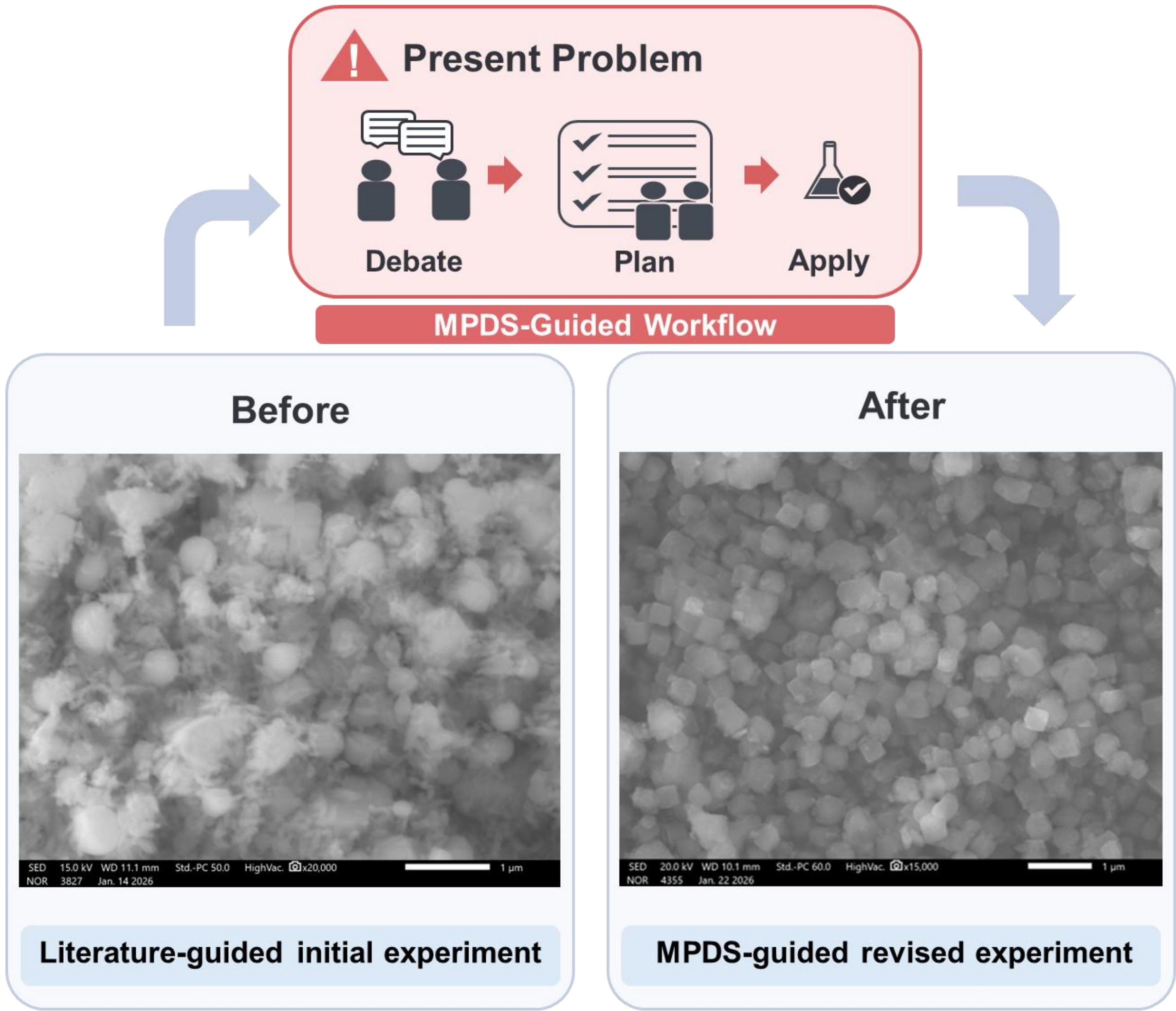

**Figure 7 | Before-and-After SEM Comparison Following MPDS-Guided Process Refinement.** Top, schematic of the MPDS-guided diagnostic loop, in which a present bottleneck is debated, translated into an experimental plan, and then applied in the laboratory. Bottom, representative SEM images from a Prussian blue analogue synthesis follow-up comparing a literature-guided initial experiment (Before) with the revised experiment after MPDS-guided intervention (After), showing visibly improved particle uniformity and preservation following process refinement. Scale bars, 1 μm.

## Discussion

In this study, we introduced MPDS, a framework that combines OpenAlex-based literature access, long-context LLM reasoning, and time-locked evaluation for reproducible benchmark testing. Across sodium-ion and all-solid-state battery case studies, MPDS synthesized large literature corpora, negotiated coupled engineering trade-offs through multi-persona debate, and generated literature-grounded, mechanistically plausible hypotheses for experimentally validated problem settings.

Methodologically, the value of MPDS lies less in using multiple agents than in using debate to make disagreements explicit, test weak assumptions, and work through competing constraints before reaching a final answer. Across the case studies and the IHQ evaluation, this structure appears most useful when the problem cannot be resolved from a single perspective and instead requires the joint consideration of mechanism, process, and trade-offs. The ablation results, therefore, support a cumulative interpretation: both persona induction and debate improve hypothesis quality relative to the evidence-only setting, and the best performance is achieved when the two are combined in MPDS. Within this pattern, debate appears to provide the larger structural step, while persona induction contributes an additional benefit by sharpening how evidence is framed, organized, and integrated into the final proposal. MPDS is therefore best understood not as a generic text-generation pipeline, but as a structured framework for literature-grounded hypothesis formation under coupled design constraints.

The current implementation has two primary limitations. First, MPDS is bounded by the quality of the ingested literature: erroneous or systematically biased sources can propagate through the system because MPDS is a synthesis engine rather than an experimental verification engine. Second, long-context ingestion is computationally expensive relative to sparse retrieval. Processing hundreds of abstracts per query increases token usage and may limit real-time scalability under constrained API budgets.

Future work will prioritize integrations that close the loop between hypothesis generation and verification. One direct path is to connect MPDS outputs to self-driving laboratory platforms by translating experimental plans into machine-executable protocols, enabling closed-loop cycles in which AI proposes, robots execute, and new data informs the next debate iteration [28,29]. A second priority is multimodal expansion. Much of the most consequential scientific evidence resides in figures (e.g., XRD and SEM), and vision-grounded interpretation combined with tool-augmented, action-grounded planning could improve executability and tighten process-iteration loops [30,31].

Although this study focuses on battery materials science, the underlying workflow may be adaptable to other text-intensive scientific domains. More specifically, MPDS addresses a recurring problem in literature-heavy research settings: how to synthesize conflicting constraints from a rapidly expanding body of scientific text. We suggest that combining long-context ingestion with multi-persona debate can be adapted to other data-intensive fields by adjusting persona definitions and OpenAlex retrieval queries. For example, medicinal-chemistry applications could pair structural-biology and clinical-pharmacology personas to balance efficacy and safety, whereas catalyst-design applications could pair activity-focused and resource-focused personas to balance performance against elemental scarcity. Unlike domain-specific systems that require task- or dataset-specific fine-tuning, MPDS conditions a general-purpose LLM on retrieved literature and performs multi-agent reasoning over aggregated long-context literature snapshots, providing a reusable reasoning workflow for text-centric scientific domains.

## Related works

### Multi-agent debate and reasoning in large language models

Recent advancements in LLMs have increasingly shifted from isolated, single-agent prompting towards multi-agent interactive frameworks to tackle complex reasoning and evaluation tasks [32,33,34]. Notably, multi-agent debate mechanisms have been demonstrated to surpass single-agent capabilities by mimicking human collaborative processes, serving as highly effective autonomous evaluators for generated content [32]. While some analyses argue that the reasoning enhancements of multi-agent discusxdvsdfvgsions are most pronounced primarily in the absence of explicit demonstrations [33], the interactive dialectic process inherently fosters the critical refinement of ideas. For instance, dialogic reasoning between opposing LLM agents has proven highly effective in recovering implicit premises within complex arguments. This highlights the capacity of structured debate to unveil unstated assumptions through iterative consensus-building and critical evaluation, rather than relying on repetitive generation [34]. Recent work has further extended this line of research from interaction alone to reliability-aware aggregation. One recent study formulates scientific multi-agent collaboration as inference-time consensus formation, using a confidence-weight table derived from agents' historical reliability and uncertainty to combine candidate responses, and reports improved performance on heterogeneous scientific reasoning and scientific proposal-writing benchmarks [35]. Unlike such general-purpose consensus frameworks, MPDS is tailored to literature-grounded hypothesis generation in battery materials through role-specific evidence pools, citation-aware multi-round debate, and moderator synthesis under a time-locked retrieval setting.

### Persona-driven frameworks for ideation and divergent thinking

Beyond convergent reasoning, the integration of distinct, well-defined personas into multi-agent frameworks has emerged as a powerful paradigm for tasks requiring creativity and divergent thinking. Persona-based collaboration significantly enriches brainstorming outcomes; empirical evidence suggests that specific persona pairings (e.g., cross-disciplinary experts) directly govern the semantic domain of ideas, thereby driving greater conceptual depth and cross-domain coverage [15]. Similarly, assigning diverse high-level beliefs and perspectives to autonomous agents facilitates the fluid, nonlinear development of complex argumentative narratives, ensuring both diversity and coherence in the final output [16].

## Data availability

The data supporting this study are available in the Article, its Supplementary Information, and via GitHub at https://github.com/ohgiracho/mpds-paper, including benchmark case metadata, case inputs, released model outputs, IHQ scoring artifacts, and OpenAlex retrieval metadata.

## Code availability

The code used in this study is available via GitHub at https://github.com/ohgiracho/mpds-paper, including scripts for retrieval, generation, evidence merging, and IHQ-based evaluation.

## References

[1] Hanson, M. A., Barreiro, P. G., Crosetto, P., & Brockington, D. (2024). The strain on scientific publishing. Quantitative Science Studies, 5(4), 823-843.

[2] Jones, B. F. (2009). The burden of knowledge and the "death of the Renaissance man": Is innovation getting harder? Review of Economic Studies, 76(1), 283–317.

[3] Schweitzer, S., & Brendel, J. (2021). A burden of knowledge creation in academic research: evidence from publication data. Industry and Innovation, 28(3), 283–306.

[4] Eppler, M. J., & Mengis, J. (2004). The Concept of Information Overload: A Review of Literature from Organization Science, Accounting, Marketing, MIS, and Related Disciplines. The Information Society, 20(5), 325-344.

[5] Sebastian, Y., Siew, E.-G., & Orimaye, S. O. (2017). Emerging approaches in literature-based discovery: techniques and performance review. The Knowledge Engineering Review, 32, e12.

[6] Frith, J. T., Lacey, M. J., & Ulissi, U. (2023). A non-academic perspective on the future of lithium-based batteries. Nature Communications, 14, 420.

[7] Huang, K. J., Ceder, G., & Olivetti, E. A. (2021). Manufacturing scalability implications of materials choice in inorganic solid-state batteries. Joule, 5(3), 564–580.

[8] Fortunato, S., Bergstrom, C. T., Borner, K., et al. (2018). Science of science. Science, 359(6379), eaao0185.

[9] OpenAI. (2023). GPT-4 Technical Report. arXiv:2303.08774.

[10] Wei, J., Wang, X., Schuurmans, D., Bosma, M., Ichter, B., Xia, F., Chi, E., Le, Q. V., & Zhou, D. (2022). Chain-of-Thought Prompting Elicits Reasoning in Large Language Models. Advances in Neural Information Processing Systems, 35, 24824–24837.

[11] Wang, X., Wei, J., Schuurmans, D., Le, Q. V., Chi, E. H., Narang, S., Chowdhery, A., & Zhou, D. (2023). Self-Consistency Improves Chain of Thought Reasoning in Language Models. International Conference on Learning Representations (ICLR 2023).

[12] Dang Anh-Hoang, Vu Tran, & Le-Minh Nguyen. (2025). Survey and analysis of hallucinations in large language models: attribution to prompting strategies or model behavior. Frontiers in Artificial Intelligence, 8, 1622292.

[13] Zhang, M., & Zhao, T. (2025). Citation Accuracy Challenges Posed by Large Language Models. JMIR Medical Education, 11, e72998.

[14] Du, Y., Li, S., Torralba, A., Tenenbaum, J. B., & Mordatch, I. (2024). Improving Factuality and Reasoning in Language Models through Multiagent Debate. Proceedings of the 41st International Conference on Machine Learning (ICML 2024), Proceedings of Machine Learning Research, 235, 11733–11763.

[15] Straub, N., Khan, S., Jay, K., Cabral, B., & Linde, O. (2025). Persona-based multi-agent collaboration for brainstorming. arXiv preprint arXiv:2512.04488.

[16] Hu, Z., Chan, H. P., Li, J., & Yin, Y. (2025). Debate-to-write: A persona-driven multi-agent framework for diverse argument generation. Proceedings of the 31st International Conference on Computational Linguistics, 4689-4703.

[17] Su, H., Chen, R., Tang, S., Yin, Z., Zheng, X., Li, J., Qi, B., Wu, Q., Li, H., Ouyang, W., Torr, P., Zhou, B., & Dong, N. (2025). Many heads are better than one: Improved scientific idea generation by a LLM-based multi-agent system. In Proceedings of the 63rd Annual Meeting of the Association for Computational Linguistics (Volume 1: Long Papers), 28201–28240. Association for Computational Linguistics.

[18] Ghafarollahi, A., & Buehler, M. J. (2025). SciAgents: Automating Scientific Discovery Through Bioinspired Multi-Agent Intelligent Graph Reasoning. Advanced Materials, 37(22), e2413523.

[19] Priem, J., Piwowar, H., & Orr, R. (2022). OpenAlex: A fully-open index of scholarly works, authors, venues, institutions, and concepts. 26th International Conference on Science, Technology and Innovation Indicators (STI 2022), Granada, Spain.

[20] Lewis, P., Perez, E., Piktus, A., Petroni, F., Karpukhin, V., et al. (2020). Retrieval-Augmented Generation for Knowledge-Intensive NLP Tasks. Advances in Neural Information Processing Systems, 33, 9459–9474.

[21] Zhuowan Li, Cheng Li, Mingyang Zhang, Qiaozhu Mei, and Michael Bendersky. 2024. Retrieval Augmented Generation or Long-Context LLMs? A Comprehensive Study and Hybrid Approach. In Proceedings of the 2024 Conference on Empirical Methods in Natural Language Processing: Industry Track, pages 881–893, Miami, Florida, US. Association for Computational Linguistics.

[22] Park, J.-S., & Kang, Y. C. (2017). Multicomponent (Mo, Ni) metal sulfide and selenide microspheres with empty nanovoids as anode materials for Na-ion batteries. Journal of Materials Chemistry A, 5(18), 8616-8623.

[23] Kim, M. J., Park, J. S., Lee, J. W., et al. (2025). Half-Covered Glitter-Cake AM@SE Composite: A Novel Electrode Design for High Energy Density All-Solid-State Batteries. Nano-Micro Letters, 17, 119.

[24] Liu, N. F., Lin, K., Hewitt, J., et al. (2024). Lost in the Middle: How Language Models Use Long Contexts. Transactions of the Association for Computational Linguistics, 12, 157-173.

[25] Gemini Team. (2025). Gemini 2.5: Pushing the Frontier with Advanced Reasoning, Multimodality, Long Context, and Next Generation Agentic Capabilities. Google.

[26] Slater, M. D., Kim, D., Lee, E., Johnson, C. S., & Doeff, M. M. (2013). Sodium-Ion Batteries. Advanced Functional Materials, 23(8), 947-958.

[27] Kerman, K., Luntz, A., Viswanathan, V., Chiang, Y.-M., & Chen, Z. (2017).
Review—Practical Challenges Hindering the Development of Solid State Li Ion Batteries. Journal of The Electrochemical Society, 164(7), A1731–A1744.

[28] Boiko, D. A., MacKnight, R., Kline, B., et al. (2023). Autonomous chemical research with large language models. Nature, 624, 570-578.

[29] Burger, B., Maffettone, P. M., Gusev, V. V., et al. (2020). A mobile robotic chemist. Nature, 583, 237-241.

[30] Yao, S., Zhao, J., Yu, D., Du, N., Shafran, I., Narasimhan, K., & Cao, Y. (2023). ReAct: Synergizing Reasoning and Acting in Language Models. International Conference on Learning Representations (ICLR 2023).

[31] Bran, A. M., Cox, S., Schilter, O., Baldassari, C., White, A. D., & Schwaller, P. (2024). Augmenting large language models with chemistry tools. Nature Machine Intelligence, 6, 525-535.

[32] Chan, C.-M., Chen, W., Su, Y., Yu, J., Liu, Z., Fu, J., Xue, W., & Zhang, S. (2023). ChatEval: Towards better LLM-based evaluators through multi-agent debate. arXiv preprint arXiv:2308.07201.

[33] Wang, Q., Wang, Z., Su, Y., Tong, H., & Song, Y. (2024). Rethinking the bounds of LLM reasoning: Are multi-agent discussions the key? Proceedings of the 62nd Annual Meeting of the Association for Computational Linguistics, 6106-6131.

[34] Ku, H. B., Shin, J., Lee, H. J., Na, S., & Jeon, I. (2025). Multi-agent LLM debate unveils the premise left unsaid. Proceedings of the 12th Argument Mining Workshop, 58-73.

[35] Yao, Y., Dong, J., Yang, Y., Li, J., & Du, Y. (2026). Roundtable Policy: Confidence-Weighted-Consensus Aggregation Improves Multi-Agent-System Reasoning. arXiv preprint arXiv:2509.16839.

## Supporting information for:

# Multi-Persona Debate System for Automated Scientific Hypothesis Generation

Q. What multicomponent Mo/Ni chalcogenide microsphere design would best preserve sodium-ion accessibility, electronic continuity, and structural buffering without wasting packing density or process scalability?

A: Hollow microspheres buffer volume change, but they still leave poor conductivity and fragile shells. I prefer in-situ Mo/Ni chalcogenide growth on an rGO scaffold so the network stays conductive and stable during cycling.

B: Your multi-step graphene-based route may work in the lab, but it is hard to scale. A spray-dried carbon-matrix microsphere offers a more practical and manufacturable path.

A: High packing density means little if electrolyte cannot access the particle interior. A flower-like Mo/Ni chalcogenide-rGO structure gives open ion pathways and short diffusion lengths.

B: Spray drying does not have to make dense, solid spheres. It can also form porous hierarchical microspheres while giving much better throughput and industrial scalability.

A: A scalable route is not enough if it still leaves dead volume and slow transport paths. A self-assembled graphene/MoNi-chalcogenide heterostructure keeps pores open, contact intimate, and buffering effective.

B: The real test is electrode-level performance, density, and process robustness. A spray-dried hierarchical microsphere is more controllable because it couples internal porosity with spherical packing and scalable processing.

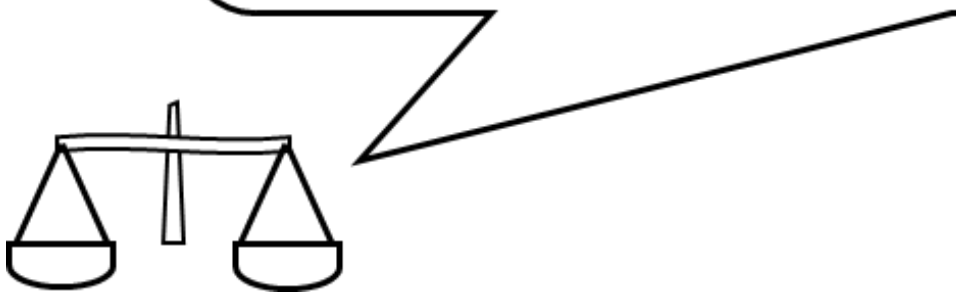

**Supplementary Figure 1 | Case Study 1 Debate Progression from Round 1 to Final Synthesis.** The sequence is shown over four panels/pages corresponding to Round 1, Round 2, Round 3, and the final synthesis. The sodium-ion-anode and particle-engineering personas initially disagree over whether graphene-rich heterostructures or spray-dried microspheres provide the better route for scalable sodium-ion-battery anodes, but the exchange progressively reframes the task as a coupled optimization of volume buffering, ion/electron transport, tap density, and manufacturability. The final synthesis converges on a graphene-reinforced nano-in-micro Mo/Ni chalcogenide microsphere fabricated by scalable spray drying.

Q. Within a fixed polycrystalline NCM811/LPSCl/existing-conductive-additive composite prepared by mechanofusion, how should limited LPSCl be geometrically allocated between particle-surface coverage and interstitial filling to best preserve ionic continuity, electronic accessibility, and dense packing at >85 wt% active material?

A: The main bottleneck is microstructural percolation at high active-material loading. A uniform LPSCl shell would electronically isolate NCM811, so we should first build a hierarchical electronic network with CNTs for local contact and CNFs for long-range conduction, then use bimodal LPSCl to create dense ionic pathways without forming a fully insulating shell.

B: That diagnosis is incomplete because the dominant failure is interfacial chemical instability, not packing alone. A stable coated NCM811 surface, a lower-reactivity CNF network, and a compatible slurry-cast composite are a more durable and scalable route.

A: Chemical stability matters, but pre-coating every particle and relying on slurry processing adds brittle layers, solvent risk, and poor scalability. A solvent-free dry-film cathode with a dual CNT/CNF network, bimodal LPSCl, and an in-situ $Li_3BO_3$-$Li_3PO_4$ buffer is the more integrated solution.

B: That dry-process strategy is too uncontrolled because in-situ buffering and dry mixing do not guarantee uniform interfaces or homogeneous dispersion. A slurry route with a designed coating, a compatible binder/solvent system, and a single optimized CNF network offers better microstructural control and interfacial precision.

A: Solvent handling remains the bigger manufacturing and chemistry risk, and brittle pre-coatings can fail during calendering. A dry-film route that avoids solvents, forms the buffer after processing, and combines CNT local contact with CNF long-range conduction is the more pragmatic path.

B: High-pressure dry processing risks cracking NCM811 and depends on speculative in-situ chemistry, whereas slurry casting enables better homogeneity and more controlled interfaces. A coated NCM811 cathode with a stable CNF network and bimodal LPSCl is therefore the more reliable engineering route.

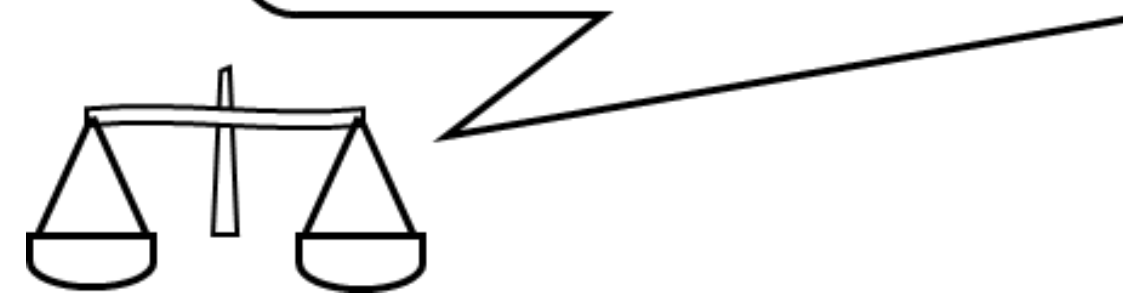

**Supplementary Figure 2 | Case Study 2 Debate Progression from Round 1 to Final Synthesis.** The sequence is shown over four panels/pages corresponding to Round 1, Round 2, Round 3, and the final synthesis. A sulfide-electrolyte perspective and a manufacturing-oriented perspective debate whether the dominant bottleneck in high-loading NCM811–LPSCl composite cathodes is interfacial instability, microstructural percolation, or process robustness. Across the three rounds, the discussion shifts from a binary slurry-versus-dry-processing dispute to a broader allocation problem spanning nano-scale surface coverage, micron-scale filler placement, conductive-network design, and manufacturability. The final synthesis endorses a stability-first hybrid route comprising pre-coated NCM811 secondary microspheres, a single low-reactivity CNF network, bimodal LPSCl, and dual-path manufacturing evaluation with slurry processing as the primary route and solvent-free dry processing as a secondary benchmark.

Q. How should vanadium nitride be nanoconfined within 3D porous reduced-graphene-oxide microspheres to raise aqueous zinc-ion cathode capacity without sacrificing reversibility, conductivity, or structural accessibility?

A: The key is not just putting VN inside rGO, but creating a defect-tolerant, intimately anchored VN/rGO composite that can activate for $Zn^{2+}$ storage. A co-assembled hierarchical structure is needed so ion access, conductivity, and mechanical stability develop together.

B: Activation chemistry matters, but morphology and process control matter just as much for a practical cathode powder. A spray-derived hierarchical microsphere can engineer porosity, packing density, and scalability better than a loosely defined self-assembly route.

A: A large hollow core wastes volume and can leave a fragile structure after nitridation. A dense but hierarchically porous VN/rGO microsphere should be built without sacrificial complexity so we keep both access and volumetric performance.

B: Simple emulsion self-assembly does not guarantee the controlled internal porosity you need. Template-assisted spray pyrolysis gives a more reproducible way to create distributed pores, rapid transport paths, and robust spherical powders.

A: Over-engineered templating can create contamination and collapse problems during harsh ammonolysis. The synthesis still has to preserve intimate VN anchoring and stable electronic pathways, not just draw a clever microsphere shape.

B: A monolithic carbon-fiber scaffold solves anchoring but fails on powder processability and volumetric density. The stronger compromise is a powder-form microsphere route that keeps hierarchical porosity while remaining compatible with scalable electrode fabrication.

The converged route is a template-free spray-pyrolyzed hierarchical VN/rGO microsphere followed by controlled ammonolysis. This keeps the powder dense and processable while preserving the internal porosity, confinement, and transport pathways needed for fast Zn-ion storage and long-cycle stability.

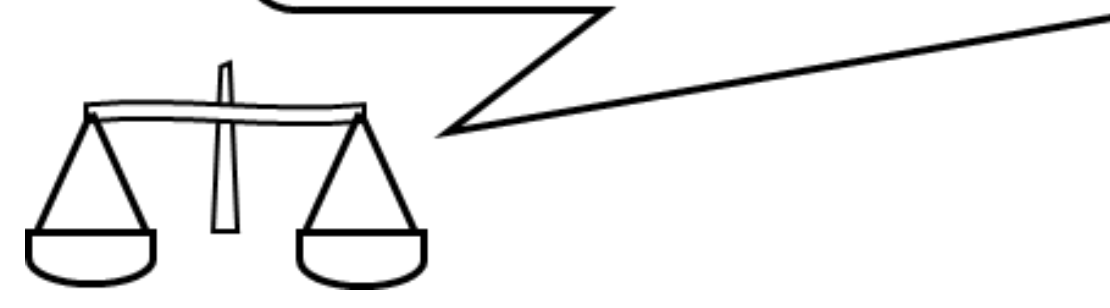

**Supplementary Figure 3 | Representative High-Scoring Debate Case from Round 1 to Final Synthesis (Case 6).** The sequence is shown over four panels/pages corresponding to Round 1, Round 2, Round 3, and the final synthesis. Two personas debate whether Zn-ion cathode design should prioritize intimate VN/rGO anchoring through self-assembly or scalable powder formation through hierarchical microspheres. The exchange progressively reconciles mechanistic activation, internal porosity, volumetric efficiency, and manufacturability. The final synthesis proposes a template-free spray-pyrolyzed hierarchical VN/rGO microsphere followed by controlled ammonolysis, preserving transport pathways and confinement while improving density and processability.

Dimension 1. Idea Novelty (0–5)

0 = Entirely conventional response with no meaningful novelty.
1 = A minor variation on a familiar idea.
2 = Some new combination or modification is present, but the idea remains close to standard practice.
3 = The response proposes a meaningfully different solution or design logic.
4 = The response introduces a clearly non-obvious idea or reconfiguration that is not simply a routine extension.
5 = The response proposes a genuinely distinctive and conceptually strong idea that would likely be recognized as notably original within the task context.

Dimension 2. Mechanistic Originality (0–5)

0 = No mechanistic reasoning is provided.
1 = Familiar mechanism-level logic is repeated without meaningful change.
2 = Known mechanisms are combined in a modest way.
3 = A more original causal chain, interaction logic, or structure–process–performance linkage is proposed.
4 = The response presents a clearly differentiated mechanism-level rationale that goes beyond routine extension.
5 = The response offers a novel mechanism-centered proposal that also leads to distinctive experimental implications or observable signatures.

Dimension 3. Trade-off Reframing (0–5)

0 = No trade-off reframing is present.
1 = The response acknowledges an existing trade-off but does not alter the reasoning around it.
2 = The response attempts to soften or partially manage a known trade-off.
3 = The response proposes a new balance point, a new optimization route, or a partial decoupling strategy.
4 = The response redefines the trade-off structure in a meaningful way, such as by introducing a new design lever or alternative route.
5 = The response transforms the problem framing so that the original trade-off is substantially bypassed, redistributed, or strategically restructured.

Dimension 4. Cross-Perspective Integration (0–5)

0 = The response reflects only one perspective.
1 = A second perspective is mentioned, but not integrated.
2 = Multiple perspectives are present, but they remain parallel rather than synthesized.
3 = The final proposal reflects meaningful integration of different viewpoints or priorities.
4 = The integration changes the structure of the proposal and produces a more nuanced or hybrid solution.
5 = The response produces a solution that clearly depends on combining perspectives that would not likely generate the same outcome independently.

**Supplementary Figure 4 | Universal IHQ Scoring Rubric.** The rubric evaluates generated hypotheses along four dimensions—Idea Novelty, Mechanistic Originality, Trade-off Reframing, and Cross-Perspective Integration—each scored on a 0–5 scale, for a total of 20 points. Used for blinded evaluation across the matched benchmark set, it operationalizes within-study comparison by distinguishing conventional answers from proposals that are mechanistically differentiated, strategically reframed, and integrative across multiple design perspectives.

| Case ID | Source/Reference Paper Full Citation | Short Task Description |
|---|---|---|
| 1 | Park, J.-S.; Kang, Y. C. Multicomponent (Mo, Ni) metal sulfide and selenide microspheres with empty nanovoids as anode materials for Na-ion batteries. Journal of Materials Chemistry A 2017, 5(18), 8616-8623. | What multicomponent Mo/Ni chalcogenide microsphere design would best preserve sodium-ion accessibility, electronic continuity, and structural buffering without wasting packing density or process scalability? |
| 2 | Kim, M. J.; Park, J.-S.; Lee, J. W.; Wang, S. E.; Yoon, D.; Lee, J. D.; Kim, J. H.; Song, T.; Li, J.; Kang, Y. C.; Jung, D. S. Half-Covered 'Glitter-Cake' AM@SE Composite: | How should a high-loading ASSB cathode composite be structured so that NCM811, LPSCl shell thickness, and small solid-electrolyte fillers jointly preserve ionic |
|  | A Novel Electrode Design for High Energy Density All-Solid-State Batteries. Nano-Micro Letters 2025, 17(1), 119. | percolation, electronic access, and packing density? |
| 3 | Park, J.-S.; Cho, J. S.; Kang, Y. C. Scalable synthesis of $NiMoO_4$ microspheres with numerous empty nanovoids as an advanced anode material for Li-ion batteries. Journal of Power Sources 2018, 379, 278-287. | How should $NiMoO_4$ microspheres with internal nanovoids be designed and processed so that void buffering, conductivity, and tap density remain balanced for durable high-rate Li-ion storage? |
| 4 | Park, J.-S.; Yang, S.; Kang, Y. C. Boosting the Electrochemical Performance of $V_2O_3$ by Anchoring on Carbon Nanotube Microspheres with Macrovoids for Ultrafast and Long-Life Aqueous Zinc-Ion Batteries. Small Methods 2021, 5(9), 2100578. | How should $V_2O_3$ be anchored on carbon nanotube microspheres with macrovoids to accelerate $Zn^{2+}$ transport while preserving structural stability and electrical continuity in aqueous ZIB cathodes? |
| 5 | Park, J.-S.; Jeong, S. Y.; Jeon, K. M.; Kang, Y. C.; Cho, J. S. Iron diselenide combined with hollow graphitic carbon nanospheres as a high-performance anode material for sodium-ion batteries. Chemical Engineering Journal 2018, 339, 97-107. | What $FeSe_2$/carbon architecture would best balance sodium storage capacity, volume-change buffering, and electrical wiring without collapsing the hollow carbon host? |
| 6 | Park, J.-S.; Wang, S. E.; Jung, D. S.; Lee, J.-K.; Kang, Y. C. Nanoconfined vanadium nitride in 3D porous reduced graphene oxide microspheres as high-capacity cathode for aqueous zinc-ion batteries. Chemical Engineering Journal 2022, 446, 137266. | How should vanadium nitride be nanoconfined within 3D porous reduced-graphene-oxide microspheres to raise aqueous zinc-ion cathode capacity without sacrificing reversibility, conductivity, or structural accessibility? |
| 7 | Park, J.-S.; Kang, Y. C. Uniquely structured Sb nanoparticle-embedded carbon/reduced graphene oxide composite shell with empty voids for high performance sodium-ion storage. Chemical Engineering Journal 2019, 373, 227-237. | What Sb/carbon/reduced-graphene-oxide shell architecture would best buffer alloying strain and preserve conductivity for high-rate sodium-ion storage without creating dead void space? |
| 8 | Park, J.-S.; Yang, S.; Kang, Y. C. Prussian blue analogue nanocubes with hollow interior and porous walls encapsulated within reduced graphene oxide nanosheets and their sodium-ion storage performances. Chemical Engineering Journal 2020, 393, 124606. | How should PBA-derived hollow sulfide nanocubes with porous walls be integrated with reduced graphene oxide to improve sodium-ion transport and structural robustness without blocking electrolyte access? |
| 9 | Park, J.-S.; Hong, J. H.; Yang, S. H.; Kang, Y. C. The conversion reaction mechanism of bimetallic Ni-Fe hydroxycarbonate and its encapsulation in carbon nanospheres for achieving excellent Li-ion storage | How should Ni-Fe hydroxycarbonate be encapsulated in carbon nanospheres so that conversion-reaction activity, electrical buffering, and structural stability remain balanced during Li-ion cycling? |
|  | performance. Journal of Materials Chemistry A 2020, 8(24), 12124-12133. |  |
| 10 | Park, J.-S.; Kim, J. H.; Kang, Y. C. Synthesis of carbonaceous/carbon-free nanofibers consisted of $Co_3V_2O_8$ nanocrystals for lithium-ion battery anode with ultralong cycle life. Electrochimica Acta 2019, 313, 48-58. | What $Co_3V_2O_8$ nanofiber architecture would best balance active-material loading, electronic conductivity, and mechanical stability for long-life lithium-ion storage? |
| 11 | Park, J.-S.; Cho, J. S.; Kang, Y. C. Nickel vanadate microspheres with numerous nanocavities synthesized by spray drying process as an anode material for Li-ion batteries. Journal of Alloys and Compounds 2019, 780, 326-333. | How should nickel vanadate microspheres with nanocavities be engineered to shorten diffusion lengths without sacrificing packing density and mechanical robustness? |
| 12 | Park, J.-S.; Lee, A.; Park, G. D.; Kang, Y. C. Synthesis of MnSe@C yolk-shell nanospheres via a water vapor-assisted strategy for use as anode in sodium-ion batteries. International Journal of Energy Research 2022, 46(3), 2500-2511. | What MnSe@C yolk-shell design would best manage sodiation-induced expansion, selenium retention, and shell conductivity without overconstraining ion access? |
| 13 | Park, J.-S.; Park, G. D.; Kang, Y. C. Exploration of cobalt selenite-carbon composite porous nanofibers as anode for sodium-ion batteries and unveiling their conversion reaction mechanism. Journal of Materials Science & Technology 2021, 89, 24-35. | How should cobalt selenite-carbon porous nanofibers be designed to manage conversion strain and conductivity loss in sodium-ion anodes while preserving fast transport? |
| 14 | Park, J.-S.; Cho, J. S.; Kim, J. H.; Choi, Y. J.; Kang, Y. C. Electrochemical properties of micron-sized $Co_3O_4$ hollow powders consisting of size controlled hollow nanospheres. Journal of Alloys and Compounds 2016, 689, 554-563. | What hollow $Co_3O_4$ secondary structure would best balance nanoscale diffusion benefits with microscale density and cycling stability in Li-ion anodes? |
| 15 | Park, S.-K.; Park, J.-S.; Kang, Y. C. Selenium-infiltrated metal-organic framework-derived porous carbon nanofibers comprising interconnected bimodal pores for Li-Se batteries with high capacity and rate performance. Journal of Materials Chemistry A 2018, 6(3), 1028-1036. | How should selenium be hosted in MOF-derived bimodally porous carbon nanofibers to balance Se confinement, ion transport, and electronic percolation in Li-Se cathodes? |
| 16 | Park, S.-K.; Park, J.-S.; Kang, Y. C. Metal-Organic-Framework-Derived N-Doped Hierarchically Porous Carbon Polyhedrons Anchored on Crumpled Graphene Balls as | How should MOF-derived porous carbon polyhedrons and crumpled graphene balls be combined to host selenium without sacrificing |
| | Efficient Selenium Hosts for High-Performance Lithium-Selenium Batteries. ACS Applied Materials & Interfaces 2018, 10(19), 16531-16540. | electrolyte access, conductivity, or structural integrity? |
| 17 | Park, S.-K.; Park, J.-S.; Kang, Y. C. Trimodally porous N-doped carbon frameworks with an interconnected pore structure as selenium immobilizers for high-performance Li-Se batteries. Materials Characterization 2019, 151, 590-601. | What trimodally porous N-doped carbon framework would best immobilize selenium while preserving electrolyte penetration and electron transport in Li-Se cathodes? |
| 18 | Park, J.-S.; Yang, S. Y.; Lee, J.-K.; Kang, Y. C. A novel strategy for encapsulating metal sulfide nanoparticles inside hollow carbon nanosphere-aggregated microspheres for efficient potassium ion storage. Journal of Materials Chemistry A 2022, 10(34), 17790-17800. | How should metal sulfide nanoparticles be encapsulated inside hollow carbon nanosphere-aggregated microspheres to stabilize potassium-ion storage without choking transport or wasting density? |
| 19 | Cho, J. S.; Park, J.-S.; Jeon, K. M.; Kang, Y. C. 1-D nanostructure comprising porous $Fe_2O_3$/Se composite nanorods with numerous nanovoids, and their electrochemical properties for use in lithium-ion batteries. Journal of Materials Chemistry A 2017, 5(21), 10632-10639. | How should porous $Fe_2O_3$/Se composite nanorods with internal nanovoids be engineered so that conversion capacity, strain buffering, and one-dimensional charge transport stay balanced? |
| 20 | Yang, S. Y.; Park, J.-S.; Koo, H. Y.; Kang, Y. C. Integration of Highly Graphitic Three-Dimensionally Ordered Macroporous Carbon Microspheres with Hollow Metal Oxide Nanospheres for Ultrafast and Durable Lithium-Ion Storage. International Journal of Energy Research 2023, 2023, 1-23. | How should highly graphitic 3DOM carbon microspheres be integrated with hollow metal oxide nanospheres to achieve fast lithium storage without sacrificing structural integrity or volumetric efficiency? |
| 21 | Park, J.-S.; Oh, Y. J.; Kim, J. H.; Kang, Y. C. Porous nanofibers comprised of hollow $SnO_2$ nanoplate building blocks for high-performance lithium ion battery anode. Materials Characterization 2020, 161, 110099. | How should hollow $SnO_2$ nanoplate-based porous nanofibers be structured to balance lithium-ion diffusion, mechanical buffering, and electrical continuity? |
| 22 | Oh, S. H.; Park, J.-S.; Jo, M. S.; Kang, Y. C.; Cho, J. S. Design and synthesis of tube-in-tube structured NiO nanobelts with superior electrochemical properties for lithium-ion storage. Chemical Engineering Journal 2018, 347, 889-899. | What tube-in-tube NiO nanobelt design would best relieve volume change while keeping ion transport pathways short and the one-dimensional framework electrically continuous? |
| 23 | Lee, J. S.; Park, J.-S.; Baek, K. W.; Saroha, R.; Yang, S. H.; Kang, Y. C.; Cho, J. S. Coral-like porous microspheres comprising polydopamine-derived N-doped C-coated $MoSe_2$ nanosheets composited with graphitic carbon as anodes for high-rate sodium- and potassium-ion batteries. Chemical Engineering Journal 2023, 456, 141118. | How should polydopamine-derived N-doped carbon-coated $MoSe_2$ nanosheets be assembled into coral-like porous microspheres to stabilize fast Na/K storage without losing conductivity or structural access? |
| 24 | Kim, J. K.; Park, S.-K.; Park, J.-S.; Kang, Y. C. Uniquely structured composite microspheres of metal sulfides and carbon with cubic nanorooms for highly efficient anode materials for sodium-ion batteries. Journal of Materials Chemistry A 2019, 7(6), 2636-2645. | What cubic-nanoroom metal sulfide/carbon microsphere design would best buffer sodium-driven strain while preserving conductive contact and scalable particle packing? |
| 25 | Cho, J. S.; Park, J.-S.; Kang, Y. C. Preparation of Hollow $Fe_2O_3$ Nanorods and Nanospheres by Nanoscale Kirkendall Diffusion, and Their Electrochemical Properties for Use in Lithium-Ion Batteries. Scientific Reports 2016, 6(1), 38933. | How should $Fe_2O_3$ hollow nanorods and nanospheres be chosen and engineered through Kirkendall diffusion to balance diffusion length, structural robustness, and electrode density? |
| 26 | Park, G. D.; Hong, J. H.; Jung, D. S.; Lee, J.-H.; Kang, Y. C. Unique structured microspheres with multishells comprising graphitic carbon-coated $Fe_3O_4$ hollow nanopowders as anode materials for high-performance Li-ion batteries. Journal of Materials Chemistry A 2019, 7(26), 15766-15773. | How should multishell $Fe_3O_4$ microspheres built from graphitic-carbon-coated hollow nanopowders be configured to improve long-cycle Li storage without excessive transport resistance or low tap density? |
| 27 | Park, G. D.; Park, J.-S.; Kim, J. K.; Kang, Y. C. Metal sulfoselenide solid solution embedded in porous hollow carbon nanospheres as effective anode material for potassium-ion batteries with long cycle life and enhanced rate performance. Chemical Engineering Journal 2022, 428, 131051. | How should metal sulfoselenide solid-solution nanoparticles and hollow carbon nanospheres be co-designed to exploit hetero-anion effects without destabilizing the anode microstructure? |
| 28 | Park, B.-I.; Park, J.-S.; Yu, S.; Cho, S.-H.; Byun, J. Y.; Oh, J.; Lee, S. Y. Hollow/porous-walled $SnO_2$ via nanoscale Kirkendall diffusion with irregular particles. Acta Materialia 2020, 186, 20-28. | How should irregular $SnO_2$ particles be transformed into hollow/porous-walled architectures through Kirkendall diffusion to improve cycling stability without losing density and connectivity? |
| 29 | Jo, S.; Kim, N.; Park, J.-S.; Chae, M. S.; Park, J. K.; Yoo, C.-Y. Practical guidelines for reliable electrochemical testing of cathodes in aqueous zinc-ion batteries: The case study of amorphous and crystalline | What testing protocol and cathode evaluation framework would most reliably distinguish intrinsic alpha-$MnO_2$ behavior from electrolyte-, counter-electrode-, and cell-configuration artifacts in aqueous ZIBs? |

| | alpha-$MnO_2$. Journal of Power Sources 2025, 647, 237325. | |
|---|---|---|
| 30 | Jeong, S.; Kim, K. B.; Jeon, M.; Kim, J. Y.; Lee, J. H.; Jang, S. Y.; Jeon, D. W.; Park, J. M.; Nahm, Y. W.; Yoo, C.-Y.; Kim, D.; Lee, J.-H.; Cho, S. B.; Kang, Y. C.; Park, J.-S. Investigation on the effect of metal incorporation in metal selenide nanoparticles and their encapsulation in hollow carbon nanospheres on improving the sodium-ion battery performance. Chemical Engineering Journal 2025, 521, 167019. | How should metal incorporation and hollow-carbon encapsulation be combined in metal selenide nanoparticles to improve sodium-ion storage without creating unnecessary mass or transport penalties? |

**Supplementary Table 1 | Benchmark case list for the 30-case Integrative Hypothesis Quality (IHQ) evaluation.** List of the 30 matched battery-materials benchmark cases used in the blinded IHQ evaluation, together with their source reference papers and time-lock settings. This table defines the held-out problem set used for controlled cross-system comparison under fixed temporal cutoffs.